\documentclass{article}

\usepackage[numbers,sort&compress]{natbib}
\usepackage[preprint]{neurips_2024}

\usepackage[utf8]{inputenc} 
\usepackage[T1]{fontenc}    
\usepackage{hyperref}       
\usepackage{url}            
\usepackage{booktabs}       
\usepackage{amsfonts}       
\usepackage{nicefrac}       
\usepackage{microtype}      
\usepackage{xcolor}         
\usepackage{graphicx}
\usepackage{array}
\usepackage{amsthm,amsmath,amssymb}

\usepackage{algorithm}
\usepackage[noend]{algpseudocode}
\usepackage{lipsum}
\usepackage[export]{adjustbox}

\title{Q*: Improving Multi-step Reasoning for LLMs with Deliberative Planning}

\author{%
    Chaojie Wang$^{1}$\footnotemark[1]\thanks{The first two authors contributed equally, and the order of authors was determined by a coin toss.} \quad
    Yanchen Deng$^{2}$\footnotemark[1] \quad 
    Zhiyi Lyu$^{2}$ \quad
    Liang Zeng$^{1}$ \quad
    Jujie He$^{1}$ \quad \\\\
    \textbf{Shuicheng Yan$^{1}$} \quad
    \textbf{Bo An$^{1}$$^{2}$}\\\\
  {$^1$Skywork AI \quad
  $^2$Nanyang Technological University}\\
}

\begin{document}

\maketitle

\begin{abstract}




Large Language Models (LLMs) have demonstrated impressive capability in many natural language tasks. However, the auto-regressive generation process makes LLMs prone to produce errors, hallucinations and inconsistent statements when performing multi-step reasoning. 
In this paper, by casting multi-step reasoning of LLMs as a heuristic search problem, we aim to alleviate the pathology by introducing Q*, a general, versatile and agile framework for guiding LLMs decoding process with deliberative planning. By learning a plug-and-play Q-value model as heuristic function for estimating expected future rewards, our Q* can effectively guide LLMs to select the most promising next reasoning step without fine-tuning LLMs for the current task, which avoids the significant computational overhead and potential risk of performance degeneration on other tasks. Extensive experiments on GSM8K, MATH and MBPP demonstrate the superiority of our method, contributing to improving the reasoning performance of existing open-source LLMs.

\end{abstract}

\section{Introduction}
Large Language Models (LLMs) have exhibited impressive capabilities in solving various reasoning tasks encoded in natural languages, including math word problems \cite{ahn2024large,cobbe2021training,hendrycks2021measuring,wang2023math,yu2023metamath,luo2023wizardmath}, code generation \cite{luo2023wizardcoder,roziere2023code,codegemma_2024} and planning \cite{xie2024travelplanner,liu2023llm+,guan2023leveraging}. Unfortunately, even the most advanced LLMs still face significant challenges and are prone to introduce errors, hallucinations and inconsistent statements as the number of reasoning steps grows due to their auto-regressive nature \cite{valmeekamMSK23,stechly2024chain}. In fact, the auto-regressive generation process of LLMs can be characterized by ``System 1" \cite{daniel2011thinking}, a mode of thought which is fast, instinctive but less accurate. Most of recent works focus on improving LLMs' ``System 1'' capability by (1) constructing sophisticated prompts with extensive expertise to trigger the potential capacities of LLMs without modifying their parameters \cite{wei2022chain,wang2022self,fu2022complexity,zhou2022least}, (2) fine-tuning LLMs with massive task-specific corpus at the price of significant computational burdens and the potential risk of performance degeneration on other tasks \cite{yu2023metamath,luo2023wizardmath,azerbayev2023llemma,yue2023mammoth}, or (3) training reward models to rank the candidate responses \cite{lightman2023let,uesato2022solving,wang2023math,khalifa2023discriminator}.

On the other hand, solving complex reasoning problems requires more in-depth, deliberative and logical thinking steps, \emph{i.e.}, the ``System 2" mode \cite{daniel2011thinking}. Taking solving math word problems as an example, any incorrect intermediate reasoning step (\emph{e.g.}, calculation errors, mis-interpretations) can potentially lead to incorrect final answers. Prior attempts \cite{yaoYZS00N23,feng2023alphazero,hao2023reasoning,zhuang2023toolchain} for enhancing ``System 2'' reasoning capability includes performing deliberation with basic tree search algorithms (\emph{e.g.}, BFS or DFS), Monte Carlo Tree Search (MCTS) \cite{browne2012survey}, and A* \cite{hart1968formal}. Nonetheless, the utility functions used in these methods often require laborious expertise to design for each specific task, which are difficult to be extended to new scenarios. Furthermore, deliberation with MCTS would require significant number of rollouts before finding high-quality responses when solving the problems with many reasoning steps, which substantially slows down the overall decoding process.

In light of this, we propose Q*, a general, versatile and agile framework for improving the multi-step reasoning capability of LLMs with deliberative planning. Different from the existing deliberation methods, our method does not rely on domain knowledge to design the heuristic function. Besides, by leveraging plug-and-play Q-value models as heuristic function, our Q* can effectively solve various tasks via guiding LLMs to select the most promising next step without fine-tuning LLMs beforehand, which avoids the significant computational overhead and potential risk of performance degeneration in other tasks. Finally, Q* considers only one single step when performing deliberation, which is much cheaper than completing rollouts in MCTS. Specifically, the main contributions of our work are summarized as follows:

\begin{itemize}
    \item We formalize the multi-step reasoning of LLMs as a Markov Decision Process (MDP) where the state is the concatenation of input prompt and the reasoning steps generated so far, the action is the next reasoning step and the reward measures how well the task is solved.
    \item We present several general approaches to estimate the optimal Q-value of state-action pairs, \emph{i.e.}, offline reinforcement learning, the best sequence from rollouts, and completion with stronger LLMs. It is noteworthy that our methods only need the ground-truth of training problems and can be easily applied to various reasoning tasks without modification.
    \item We cast solving multi-step reasoning tasks as a heuristic search problem, where the objective is to find the most proper reasoning trace with maximum utility. Built upon A* search, our deliberation framework, Q*, leverages plug-and-play Q-value models as heuristic function and guides LLMs to select the most promising next reasoning step in best-first fashion.
    \item We conduct extensive experiments on math reasoning and code generation tasks, demonstrating that Q* can significantly improve the multi-step reasoning capability of existing open-sourced LLMs.
    
\end{itemize}






    

    
    

\section{Related Works}
\paragraph{LLM alignment.} Alignment has become an important technique to prevent the output of LLMs deviates from human's expectation. Supervised Fine-Tuning (SFT) is probably the most fundamental alignment approach that directly minimizes the cross-entropy loss between the output and ground-truth. Reinforcement learning from Human Feedback
(RLHF) \cite{ouyang2022training}, on the other hand, firstly learns a reward model (RM) from human preferences and then optimizes the SFT model with reinforcement learning algorithms to maximize the cumulative rewards from RM. Direct Preference Optimization (DPO) \cite{rafailov2023direct} aligns LLMs directly according to the ranking information from human feedback without explicitly learning RM. Recently, Aligner \cite{ji2024aligner} came out as a model-agnostic alignment method by learning to re-write LLMs' output. Compared to these methods, our Q* achieves the goal of alignment with distinct merits.   Different from SFT and Aligner, Q* does not rely on massive human annotated preference pairs which are expensive to collect; different from RLHF and DPO, our Q* does not modify the parameters of LLMs, which avoids the potential risk of performance degeneration on other tasks. 
\paragraph{Enhancing LLMs with planning.} Tree-of-thoughts (ToT) \cite{yaoYZS00N23} improves the LLMs' reasoning capability by exploring the intermediate steps towards problem solving with basic tree-search algorithms. In the same vein, A* search and MCTS are applied to serve as a planning technique to enhance the performance of LLMs when solving challenging complex reasoning problems \cite{feng2023alphazero,hao2023reasoning,zhuang2023toolchain,hazra2024saycanpay}. Unfortunately, the utility function used in these methods is either constructed from LLMs' feedback (\emph{e.g.}, \cite{yaoYZS00N23,hao2023reasoning}), which could be highly-inaccurate in complex problems, or specific to each individual task (\emph{e.g.}, \cite{zhuang2023toolchain,hazra2024saycanpay}). Moreover, planning with MCTS often requires to perform costly rollout, which can significantly slow down the overall decoding process. In contrast, our Q* solely relies on training a Q-value model to guide LLMs to select the most promising next reasoning step and the pipeline can be easily applied to various reasoning tasks without modification. Besides, we consider only a single step each time in Q*, which is much cheaper than complete rollout in MCTS-based methods. 
\paragraph{LLMs for math reasoning \& code generation.} Math reasoning and code generation require LLMs to perform multi-step reasoning on relations, quantities and logics which are inherently challenging. Current techniques include: 1) prompt engineering which triggers the potential capacities of LLMs with sophisticated prompts \cite{wei2022chain,wang2022self,fu2022complexity,zhou2022least,huang2023agentcoder,shinn2023reflexion}. However, constructing such prompt needs extensive expertise and case-by-case tuning, which is difficult to generalize to different tasks; 2) Fine-tuning LLMs with massive math/code corpus \cite{yu2023metamath,luo2023wizardmath,azerbayev2023llemma,yue2023mammoth,roziere2023code,codegemma_2024,codeqwen1.5}, which usually comes at the price of significant computational burden and may compromise the performance on other tasks; 3) training RMs/verifiers to rank the candidate solutions without providing any guidance in intermediate steps \cite{lightman2023let,uesato2022solving,wang2023math,khalifa2023discriminator}. Differently, our Q* leverages a plug-and-play Q-value model to direct the deliberation process of LLMs, which effectively provides guidance for each intermediate step without modifying the parameters of LLMs. Moreover, by casting multi-step reasoning of LLMs as a heuristic search problem, our method can be generalized to various reasoning tasks without laborious prompt engineering.

\section{Preliminary}

\subsection{Formulate the Multi-step Reasoning of LLMs as an MDP}

Taking the question $q$ as input, the answer generation process of LLMs can be broken down into multiple reasoning steps, where the final answer sequence $\mathbf{a}$ can be treated as the concatenation of these $T$ single-step reasoning steps, formulated as $\mathbf{a}=[a_1;a_2;\dots;a_T]$. Each step can be a single line or fixed number of tokens outputted by LLMs.
Under this perspective, we can conceptualize the multi-step reasoning process of LLMs as a Markov Decision Process (MDP) $\langle \mathcal{S},\mathcal{A},\mathcal{T},\mathcal{R},\gamma\rangle$, where the state $s_t \in \mathcal{S}$ denotes the concatenation of the input question and the partial reasoning trace already generated by timestep $t-1$ (\emph{i.e.}, $s_t=[q;a_1;\dots;a_{t-1}]$) with the special case $s_1=q$,  the action $a_t \in \mathcal{A}$ denotes the next reasoning step generated by LLMs taking the current state $s_t$ as input, the deterministic state transition $\mathcal{T}$ from the current state $s_t$ to the next state $s_{t+1}$ is accomplished through a simple operation of
concatenation, $\mathcal{R}$ is the reward function to measure how well the question is solved and $\gamma$ is the discount factor. The reward function is often \textit{outcome-based}. That is, it gives reward by comparing the final results with ground-truth:
\begin{equation}
\mathcal{R}(s_t,a_t) = \begin{cases}
1 &t=T\land [s_t;a_t]\;\text{matches the ground-truth}\\
0&\text{otherwise}
\end{cases},
\end{equation}
In particular, we assign a reward of 1 if the generated code passes all test cases (for code generation tasks) or the final answer matches the ground-truth (for math reasoning tasks) which is a common practise in previous studies \cite{wang2023math,lightman2023let}.
Finally, the policy $\pi_{\theta}$ is embodied by an LLM, which produces reasoning sequence conditioned on the input question:
\begin{align}
\pi_{\theta}(a_t|s_t) = \text{LLM}(a_t|s_t), ~ \pi_\theta(\mathbf{a}|q)=\prod_{t=1}^{T}\pi_\theta(a_t|s_t).\label{eq-auto-reg}
\end{align}

Given the MDP and LLM policy $\pi_\theta$, the \textit{value} of state-action pair $(s_t,a_t)$ is given by a \textit{Q-function} $Q^{\pi_\theta}(s_t,a_t)=\mathbb{E}_{\pi_\theta}\left[\sum_{t^\prime=t}^T\gamma^{T-t^\prime}\mathcal{R}(s_{t^\prime},a_{t^\prime})\right]$. The Q-function of an optimal policy $\pi^*$ is called \textit{optimal Q-function} and satisfies the Bellman optimality equation:
\begin{equation}
    Q^*(s_t,a_t)=\mathcal{R}(s_t,a_t)+\gamma\max_{a_{t+1}\in\mathcal{A}}Q^*(s_{t+1},a_{t+1}), \label{eq-opt-q}
\end{equation}
which gives the value of starting state $s_t$, taking action $a_t$ and then following the optimal policy $\pi^*$.

\subsection{A* Search}

\textbf{A*} \cite{hart1968formal} is an important heuristic search algorithm in deliberative planning \cite{bonet2001planning}, multi-agent pathfinding \cite{silver2005cooperative}, and constraint reasoning \cite{pezeshki2022and}. Originally, A* is proposed for finding the shortest path from source $s$ to goal $g$ in path planning problems. It associates each frontier vertex $n$ with a value $f(n)=g(n)+h(n)$, where $g(n)$ is the accumulated path cost from source $s$ and $h(n)$ is a heuristic value that estimates the cost of the shortest path 
from $n$ to goal $g$. The algorithm adopts a best-first search strategy, \emph{i.e.}, in each iteration it always picks the vertex with minimum $f$-value to explore until reaching the goal. When the heuristic $h(\cdot)$ is \textit{admissible} \cite{russell2016artificial}, A* guarantees to find the optimal path.





\begin{figure}
    \includegraphics[width=.49\linewidth,valign=c]{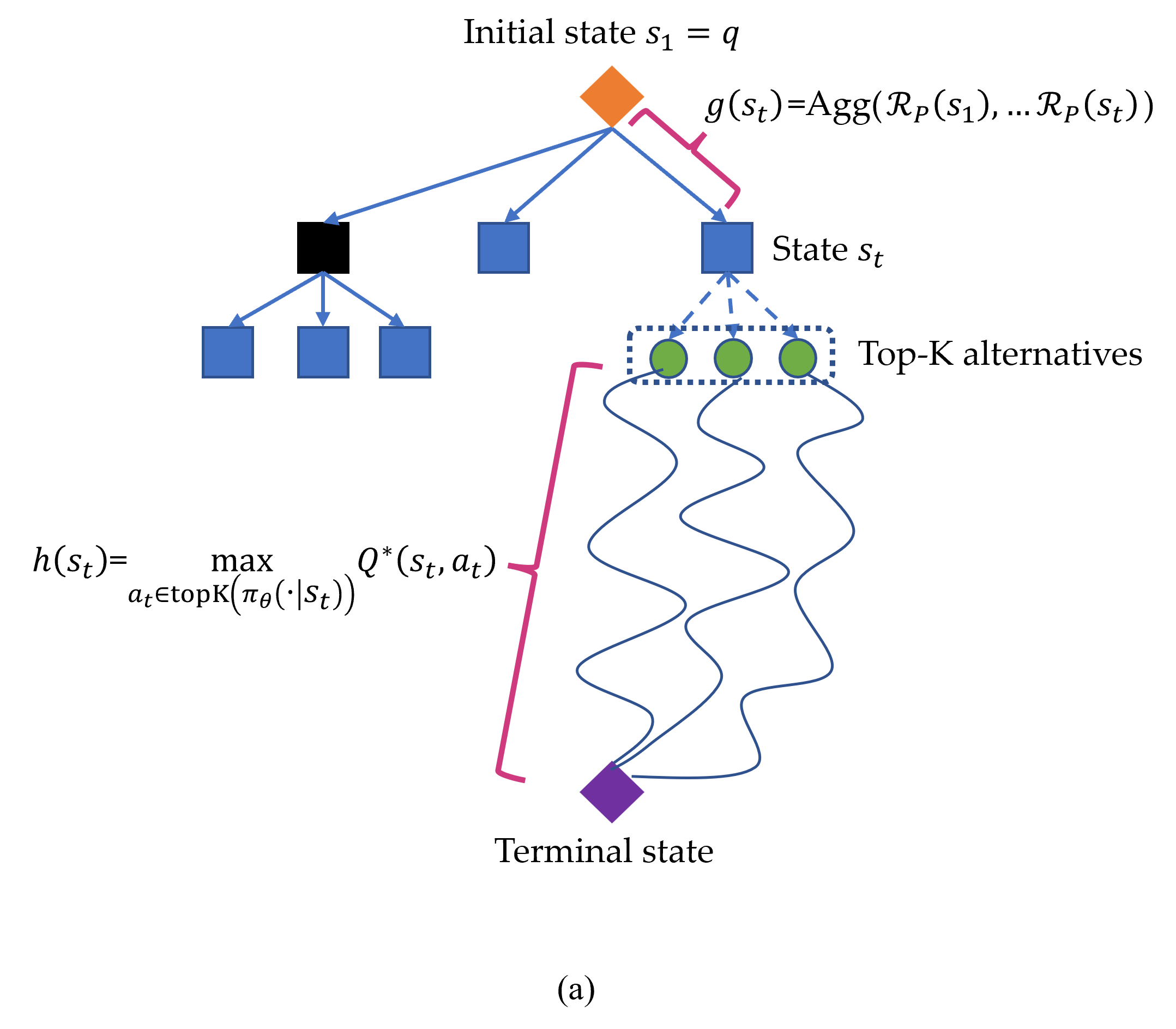}\hfill
    \includegraphics[width=.49\linewidth,valign=c]{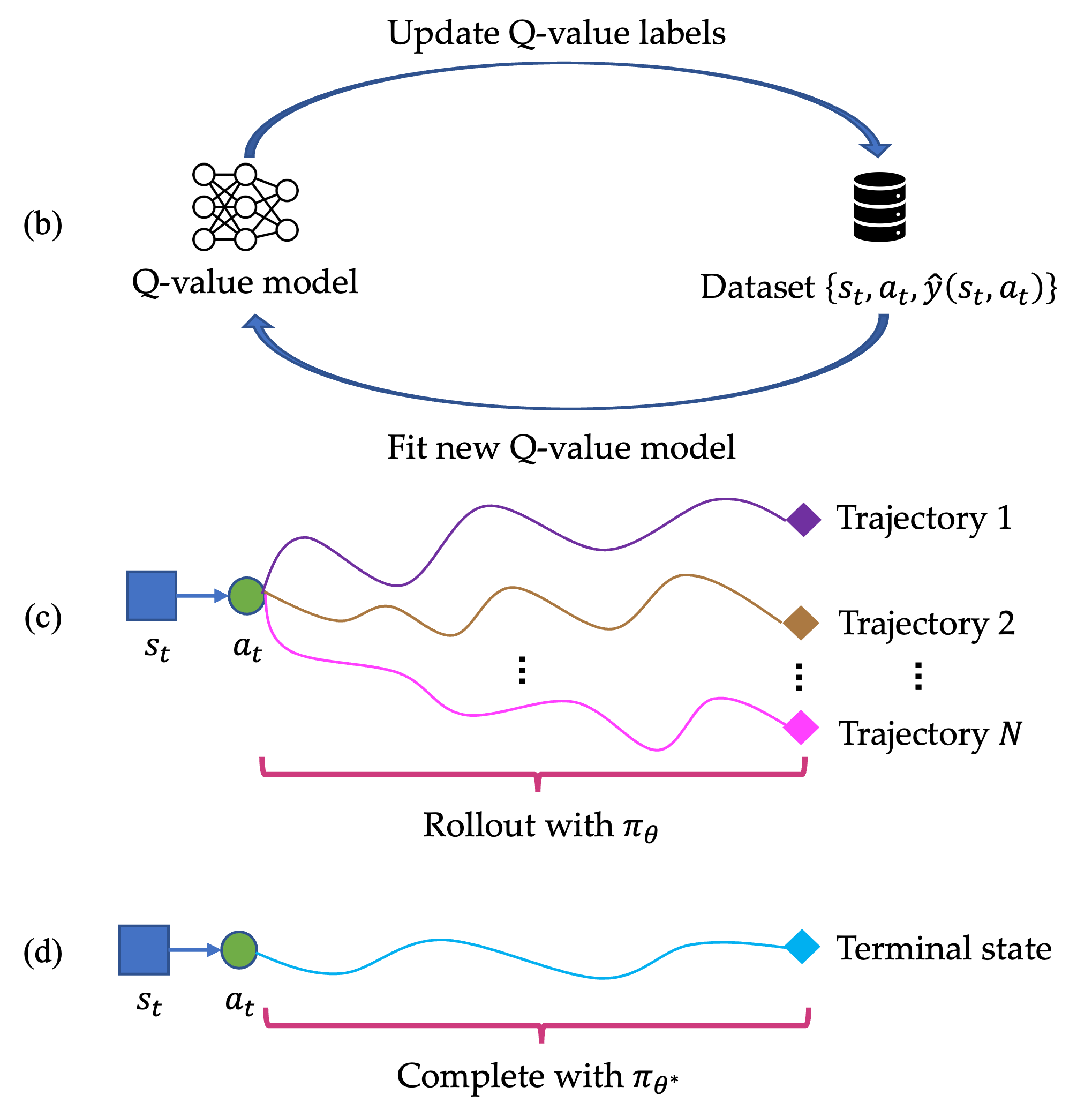}
    \caption{Overview of Q*. \textbf{(a)}: the deliberation process of Q*. Each state is associated with an $f$-value which is the weighted sum of the aggregated utility (cf.~Eq.~(\ref{eq-g})) and the heuristic value (cf.~Eq.~(\ref{eq-f-value})). \textbf{(b-d)}: estimating optimal Q-value with fitted-Q-iteration, rollout and completion with stronger LLMs.}
\end{figure}

\section{Q*: A General, Versatile and Agile Deliberation Framework for LLMs}
Most of modern LLMs generate natural languages in an auto-regressive way, \emph{i.e.}, predict the next token in a sequence given the previously generated tokens (cf. Eq.~(\ref{eq-auto-reg})). Therefore, when applied to multi-step reasoning problem, LLMs can potentially introduce errors, hallucinations and inconsistent statements in the subsequent reasoning trace if any previous step is incorrect, which may fail to solve the current problem. Indeed, given the fact that LLMs produce each token with limited computation resources, there is no way to devote more computational efforts to solve difficult problems. In short, LLMs cannot perform in-depth deliberation which is essential for solving complex multi-step reasoning problems.

We address this issue by presenting Q*, a general, versatile and agile deliberation framework based on A* to effectively guide LLMs to select the most promising next step when performing multi-step reasoning without costly fine-tuning LLMs for each task beforehand. In more detail, we cast finding the most proper reasoning sequence for a given problem as a heuristic search process, where each state $s_t$ is associated with a $f$-value estimating how much utility will be attained if we expand $s_t$:
\begin{align}
f(s_t) = g(s_t) +\lambda h(s_t),
\end{align}
where $g(s_t)$ denotes the aggregated utility from the initial state $s_1$; $h(s_t)$ is the heuristic value for measuring the probability of reaching the correct answer derived from $s_t$; $\lambda$ is a coefficient to balance the importance of $g(s_t)$ and $h(s_t)$ terms.

Specifically, we propose to use process-based reward function $\mathcal{R}_P$ that encodes the prior knowledge or preference of the reasoning task to compute the aggregated utility $g(s_t)$. That is,
\begin{equation}
    g(s_t)=\text{Agg}(\mathcal{R}_P(s_1),\dots,\mathcal{R}_P(s_i),\dots,\mathcal{R}_P(s_t)),\label{eq-g}
\end{equation}
where Agg~$\in\{\min, \max, \sum, [-1]\}$, with 
$[-1]$ standing for assigning the reward of last state as the utility, is the aggregation function to summarize the rewards in the path from $s_1$ to $s_t$, and $s_{i-1}$ is the prefix of $s_i,\; 1<i\le t$. Such process-based reward function $\mathcal{R}_P$ could be learned from human feedback \cite{lightman2023let,uesato2022solving,wu2023fine}, ground-truth \cite{wang2023math,khalifa2023discriminator}, rules, or simply be the logits of a reasoning step which reflects the confidence of the LLM. Furthermore, we use the optimal Q-value of state $s_t$ (cf. Eq.~(\ref{eq-opt-q})) as the heuristic value $h(s_t)$. In other words, the $f$-value is given by:
\begin{align}
f(s_t) = g(s_t) +\lambda \max_{a_t\in\mathcal{A}}Q^*(s_t,a_t).\label{eq-f-value}
\end{align}
Since enumerating all possible next reasoning steps is intractable, in practice we restrict the alternatives to the top-K of all step candidates returned by LLM, and thus Eq.~\eqref{eq-f-value} can be written as  $f(s_t) = g(s_t) +\lambda \max_{a_t\in\text{top-K}(\pi_\theta(\cdot|s_t))}Q^*(s_t,a_t)$.

\subsection{Estimation of Optimal Q-value}
\label{sec_q_estimation}

A critical challenge of implementing Q* is to estimate the optimal Q-value of state-action pairs (cf.~Eq.~(\ref{eq-f-value})) with a frozen LLM policy $\pi_\theta$ which could be suboptimal on the given reasoning problems. Therefore, we aim to learn a proxy Q-value model $\hat{Q}$ to approximate $Q^*$ from a dataset $D=\{q_i,\{\mathbf{a}_{i_j}\}_{j=1}^M\}_{i=1}^N$, where $q_i$ is a training problem and $\mathbf{a}_{i_j}\sim\pi_\theta(\cdot|q_i)$ is a trajectory sampled from the LLM policy $\pi_\theta$. Formally:
\begin{equation}
    \hat{Q}=\arg\min_Q\frac{1}{NMT}\sum_{i=1}^N\sum_{j=1}^M\sum_{a_t\in\mathbf{a}_{i_j}}\left(Q(s_t,a_t)-\hat{y}(s_t,a_t)\right)^2,\label{eq-q-learning}
\end{equation}
where $s_t=[q_i;a_1;\dots;a_{t-1}]$ is the partial reasoning trace by timestep $t-1$ in $\mathbf{a}_{i_j}$ and $\hat{y}(s_t,a_t)$ is the label that approximates the true optimal Q-value, specifically $Q^*(s_t,a_t)$.
 
In more detail, we effectively construct Q-value labels $\hat{y}(s_t,a_t)$ for question $q_i$ in the following ways:
\paragraph{Offline reinforcement learning.} Given the offline dataset $D$, we learn the proxy Q-value model $\hat{Q}$ using Fitted Q-iteration \cite{riedmiller2005neural}. Specifically, for each iteration $\ell$, we construct Q-value label as:
\begin{equation}
    \hat{y}_\ell(s_t,a_t) = \begin{cases}
\mathcal{R}(s_{t},a_{t}) &t=T\\
\mathcal{R}(s_{t},a_{t}) +\gamma \max_{a_{t+1}\in\text{top-K}(\pi_\theta(\cdot|s_{t+1}))} \hat{Q}_{\ell-1}(s_{t+1},a_{t+1})&\text{otherwise}
\end{cases},
\end{equation}
where $\hat{Q}_{\ell-1}$ is the proxy Q-value model learned in iteration $\ell-1$. After that, we train a new proxy model $\hat{Q}_\ell$ according to Eq.~(\ref{eq-q-learning}). Such two phases will be alternated for $L$ iterations, and we use $\hat{Q}_L$ as the proxy Q-value model when performing deliberation.

\paragraph{Learning from rollout.} Staring from the state-action pair $(s_t,a_t)$, we first perform random rollout or MCTS \cite{hao2023reasoning,feng2023alphazero} with policy $\pi_\theta$ for enough iterations, forming a trajectory pool $P$. After that, we use the best reasoning sequence with the highest accumulated rewards to construct the Q-value label:
\begin{equation}
    \hat{y}(s_t,a_t)=\mathcal{R}(s_{t},a_{t})+\max_{\substack{\tau\sim P\\(s_{t^\prime},a_{t^\prime})\in\tau}}\left[\sum_{t^\prime=t+1}^T\gamma^{T-t^\prime}\mathcal{R}(s_{t^\prime},a_{t^\prime})\right].
\end{equation}
\paragraph{Approximating optimal policy with stronger LLMs.} By exploiting the fact that the optimal Q-values are derived from an optimal policy, one can estimate the optimal Q-value of state-action pair $(s_t,a_t)$ by completing the trajectory with another stronger LLM $\pi_{\theta^*}$ (\emph{e.g.}, GPT-4) if available:
\begin{equation}
    \hat{y}(s_t,a_t)=\mathcal{R}(s_{t},a_{t})+\sum_{t^\prime=t+1}^T\gamma^{T-t^\prime}\mathcal{R}(s^*_{t^\prime},a^*_{t^\prime}),
\end{equation}
where $s^*_{t^\prime}=[s_t;a_t;a^*_{t+1};\dots;a^*_{t^\prime-1}]$ and $a^*_{t^\prime}\sim\pi_{\theta^*}(\cdot|s^*_{t^\prime})$.



\subsection{Deliberative Planning with A*}
Once obtaining the proxy Q-value model $\hat{Q}$, we can plug it to Eq.~(\ref{eq-f-value}) to compute the $f$-value of each state and perform best-first search with A*. Alg.~\ref{algo-a-star} illustrates the deliberative planning process.
\algrenewcommand\algorithmicrequire{\textbf{Input:}}
\algnewcommand\algorithmicforeach{\textbf{for each}}
\algdef{S}[FOR]{ForEach}[1]{\algorithmicforeach\ #1\ \algorithmicdo}
\begin{algorithm}
\centering
\caption{Deliberative planning for LLMs with A*} \label{algo-a-star}
\begin{algorithmic}[1]
\Require question $q$, LLM policy $\pi_\theta$, proxy Q-value model $\hat{Q}$
\State $unvisited \gets\{q\}$, $visited\gets\emptyset$
\While{$unvisited \ne\emptyset$}
\State $s\gets\arg\max_{s^\prime\in unvisited}f(s^\prime)$
\State $unvisited \gets unvisited \backslash\{s\}$, $visited\gets visited\cup\{s\}$
\If{$s$ is a terminal state}
\Return the complete trajectory $s$
\EndIf
\ForEach{$a\in\text{top-K}(\pi_\theta(\cdot|s))$}
\State $s^\prime\gets [s;a]$
\If{$s^\prime\notin visited$}
$unvisited \gets unvisited \cup\{s^\prime\}$
\EndIf
\EndFor
\EndWhile
\end{algorithmic}
\end{algorithm}

Specifically, we maintain a set for storing state candidates to be explored, denoted as $unvisited$, which initially only contains the input question $q$, and another set $visited$ to record the visited states. Each step we pick the state $s$ with the maximum $f$-value from the set $unvisited$ and expand it by querying the top-K alternatives with the LLM policy $\pi_\theta$. After that, both $visited$ and $unvisited$ sets will be updated and this process repeats until the terminal state (a complete trajectory) is reached. Finally, we extract the answer part of the terminal state $s$ as the result.

\section{Experiments}

\subsection{Experimental Settings}

\textbf{Datasets.} We evaluate the effectiveness of Q* on two math reasoning and one code generation tasks, where the dataset statistics have been summarized in Table \ref{tab_dataset_statics}. 
1) GSM8K \cite{cobbe2021training} is a dataset of grade school math problems, where the solution is given in a one-line-per-step format with an exact numerical answer in the last line;
2) MATH \cite{hendrycks2021measuring} is a dataset consisting of math problems of high school math competitions, where the solutions are given in a format that mixes latex code and natural language;
3) MBPP \cite{austin2021program} is an entry-level Python programming dataset, where the questions are coding challenges along with a test case that defines the function format.
The solutions are Python code that is excepted to pass the pre-collected test cases of each question.

\begin{table}[h]
 \centering
 \caption{Statistics of datasets.
 }
 \label{tab_data}
 {
 \begin{tabular}{c|c|c|c}
 \toprule
 Dataset  & GSM8K & MATH & MBPP\\
 \midrule 
 Domain & Math Reasoning & Math Reasoning & Code Generation\\
 Training & 5000 & 8000 & 374\\
 Testing & 1319 & 5000 & 500\\
 Average Steps & 4.5 & 11.0 & 7.0 \\
 \bottomrule
 \end{tabular}
 }
 \label{tab_dataset_statics}
\end{table}

\textbf{Implementation Details.} The implementation of Q* method mainly includes three steps: 
1) Q-value estimation. As discussed in Section~\ref{sec_q_estimation}, we propose several ways for estimating Q values and will illustrate the implementation details in the next paragraph; 
2) Utility aggregation for each step (cf. Eq.~(\ref{eq-g})).
For GSM8K dataset, we adopt a process reward model (PRM) trained on PRM800K \cite{lightman2023let} to model $\mathcal{R}_P$ to provide an intermediate signal for each reasoning step, and use $\min$ as the aggregation function;
For MATH dataset, we set $g(s_t)=0$ for all passed states $\{s_i\}_{i=1}^{t}$ in each trajectory for fairness, because PRM800K contains data samples constructed from MATH testing set and there is a potential risk of data leakage;
For MBPP dataset, we tokenize the code generated so far with function \verb|tokenize.generate_tokens| and give a penalty of -0.5 if \verb|TokenError| is raised, which is often the case that there are mismatched delimiters (e.g., parentheses, quotation marks) and invalid indention in the code. We use $[-1]$ as the aggregation function to cancel the previous penalties since the code is generated on-the-fly and mismatched delimiters may be fixed in subsequent steps.
3) A* planning. For GSM8K and MATH datasets, we treat a single line outputted by the LLM as an action, while the action in MBPP is defined as a code snippet with 24 tokens when planning. Besides, in all experiments, we set $\lambda=1$ when computing $f$-values and expand a state with $K=6$ actions at each reasoning step. Finally, following the common practice of Best-of-$N$, we perform planning to collect $N=6$ trajectories for each question, and select the one with the maximum $f$-value as the final result for evaluation.

For Q-value estimation, in our practice, we find that learning from rollout could be the most effective and robust way to collect precise Q-value labels.
Specifically, given the prompt of questions, we will firstly perform random rollout to obtain complete trajectories with LLM, denoted as $\pi_\theta$, under the setting of high temperature, \emph{e.g.}, $\tau=0.9$ for math reasoning and $\tau=0.2$ for code generation, 
and split each trajectory into a series of states according to the newline token ``$\textbackslash n$''.
Then, for each state-action pair in a trajectory, denoted as $(s_t, a_t)$, we can perform random rollout or MCTS with the same LLM to collect a pool $P$ of trajectories, and then select the best reasoning path with the highest accumulated rewards to construct the corresponding Q-value label of the current state-action pair.
Note that the reward $\mathcal{R}(s_t,a_t)$ is given as 1 only if the obtained math numerical answer is correct or the program passes all test cases, indicating that the Q value of a state-action pair can be 1 only if it has the potential to generate a trajectory containing the correct answer.




\subsection{Experimental Results}

\textbf{GSM8K.} For the comparison on GSM8K dataset, we select Llama-2-7b \cite{touvron2023llama} as our base model, whose accuracy can achieve 65.2\% after finetuning on MetaMath \cite{yu2023metamath}.
Then, we treat Llama-2-7b finetuned on MetaMath as policy $\pi_\theta$, and perform random rollout to collect Q-value labels for training Q-value model (QVM).
For utility aggregation, we train a process reward model (PRM) on PRM800K \cite{lightman2023let} to provide intermediate signal for each reasoning step.
With PRM and QVM in hand, traditional methods tend to treat either of them as a verifier to select the Best-of-$N$ trajectory or utilize them to perform PPO training of RLHF. 
As the results shown in Table~\ref{tab_result_gsm8k}, we can find that with the same PRM/QVM, using it for verification performs significantly better than using it for alignment.
Further, in the comparison of planning-based methods, we can find that with the same QVM, Q* method with constant aggregated utility can still outperform Best-of-$N$ method.
With the PRM trained on PRM800K determining whether the intermediate reasoning steps are correct, Q* method that combines PRM and QVM achieves the best performance among all methods based on the same LLM, helping Llama-2-7b surpass the performance of close-sourced ChatGPT-turbo \cite{shridhar2023art} and reaching an accuracy of 80.8\%.

\textbf{MATH.} As the results shown in Table~\ref{tab_math}, considering the weak performance of Llama-2-7b fine-tuned with MetaMath for the MATH dataset, we seek for two other stronger LLMs to evaluate the effectiveness of our Q* method.
One is Llama-2-7b fine-tuned on Synthetic Data \cite{li2024common}, which is constructed following the instruction of scaling up the SFT data, and achieves 41.9\% accuracy on MATH dataset, approaching the performance of GPT-4 \cite{bubeck2023sparks}.
The other base model is DeepSeek-Math-7b \cite{shao2024deepseekmath}, which could be the most powerful open-source 7b model for math reasoning on MATH dataset, achieving 50.8\% accuracy in our evaluation.
From the results shown in the second and third blocks of Table~\ref{tab_math}, we can find that Q* can still lead to further performance improvement compared to the Best-of-$N$ method on either of base models. Additionally, it is noteworthy that the performance of DeepSeek-Math-7b enhanced with Q* has already surpassed a series of closed-source models on the leaderboard of MATH dataset \footnote{\url{https://paperswithcode.com/sota/math-word-problem-solving-on-math}}, such as Gemini Ultra (4-shot), reaching an accuracy of 55.4\% .

\begin{table*}
\caption{Comparison of Q* and other baselines on GSM8K dataset.}
\renewcommand{\arraystretch}{1.25}
\centering
\scalebox{0.75}{
   \begin{tabular}{
   p{4.5cm}<{\centering}| p{3cm}<{\centering} | p{3cm}<{\centering} | p{3.5cm}<{\centering} | p{2.5cm}<{\centering}} 
   \toprule 
    \multicolumn{1}{c|}{Base Model} &\multicolumn{1}{c|}{SFT} &\multicolumn{1}{c|}{Alignment} &\multicolumn{1}{c|}{Verification} &\multicolumn{1}{c}{Accuracy}\\
    \midrule
    GPT-3.5 (5-shot) \cite{achiam2023gpt}   & Unknown & PPO (RM) \cite{ouyang2022training}& - & 57.1\% \\
    ChatGPT-instruct (0-shot) \cite{shridhar2023art} & Unknown & PPO (RM) \cite{ouyang2022training} & - & 71.3\% \\
    ChatGPT-turbo (0-shot) \cite{shridhar2023art} & Unknown & PPO (RM) \cite{ouyang2022training} & - & 77.7\% \\
    GPT-4 (0-shot) \cite{shridhar2023art}   & Unknown & PPO (RM) \cite{ouyang2022training} & - & 91.9\% \\
    GPT-4 (5-shot) \cite{achiam2023gpt}   & Unknown & PPO (RM) \cite{ouyang2022training} & - & \textbf{92.0}\% \\
   \midrule 
    Llama-2-7b (0-shot)   &- &-         &- &49.5\% \\
    Llama-2-7b (0-shot) &MetaMath\cite{yu2023metamath} &-         &- &65.2\% \\
    Llama-2-7b (0-shot) &MetaMath\cite{yu2023metamath} &PPO (PRM) \cite{ouyang2022training} &- & 67.2\% \\
    Llama-2-7b (0-shot) &MetaMath\cite{yu2023metamath} &PPO (QVM) \cite{ouyang2022training} &- & 67.6\% \\
    Llama-2-7b (0-shot) &MetaMath\cite{yu2023metamath} &-         &Best-of-$N$ (PRM) \cite{lightman2023let}& 72.1\%  \\
    Llama-2-7b (0-shot) &MetaMath\cite{yu2023metamath} &-         &Best-of-$N$ (QVM) \cite{lightman2023let}& 74.5\%  \\
    Llama-2-7b (0-shot) &MetaMath\cite{yu2023metamath} &-         &Q* (QVM)& 78.8\% \\
    Llama-2-7b (0-shot) &MetaMath\cite{yu2023metamath} &-         &Q* (PRM+QVM)& \textbf{80.8}\% \\
   \midrule
   \end{tabular}}
   \label{tab_result_gsm8k}
\end{table*}

\begin{table*}
\caption{Comparison of Q* and other baselines on MATH dataset.}
\renewcommand{\arraystretch}{1.25}
\centering
\scalebox{0.75}{
   \begin{tabular}{
   p{4.5cm}<{\centering}| p{3cm}<{\centering} | p{3cm}<{\centering} | p{3.5cm}<{\centering} | p{2.5cm}<{\centering}} 
   \toprule 
    \multicolumn{1}{c|}{Base Model} &\multicolumn{1}{c|}{SFT} &\multicolumn{1}{c|}{Alignment} &\multicolumn{1}{c|}{Verification} &\multicolumn{1}{c}{Accuracy}\\
    \midrule
    GPT-3.5 (0-shot) \cite{bubeck2023sparks} & Unknown & PPO (RM) \cite{ouyang2022training} & - & 23.5\% \\
    GPT-4 (0-shot) \cite{bubeck2023sparks}  & Unknown & PPO (RM) \cite{ouyang2022training} & - & 42.5\% \\
    Gemini Ultra (4-shot) \cite{team2023gemini} & Unknown & PPO (RM) \cite{ouyang2022training} & - & \textbf{53.2}\% \\
   \midrule 
    Llama-2-7b (0-shot) &- &-         &- &7.9\% \\
    Llama-2-7b (0-shot) &MetaMath\cite{yu2023metamath} &-         &- &20.0\% \\
    Llama-2-7b (0-shot) &Synthetic Data\cite{li2024common} &-         &- &41.9\% \\
    Llama-2-7b (0-shot) &Synthetic Data\cite{li2024common} &PPO (QVM) \cite{ouyang2022training} &- &42.5\% \\
    Llama-2-7b (0-shot) &Synthetic Data\cite{li2024common} &-         &Best-of-$N$ (QVM) \cite{lightman2023let} &46.8\% \\
    Llama-2-7b (0-shot) &Synthetic Data\cite{li2024common} &-         &Q* (QVM)&\textbf{49.1}\% \\
   \midrule
    DeepSeek-Math-7b (0-shot) & Unknown &PPO (QVM) \cite{ouyang2022training}          & -&50.8\% \\
   DeepSeek-Math-7b (0-shot) & Unknown &PPO (QVM) \cite{ouyang2022training}          & Best-of-$N$ (QVM) \cite{lightman2023let} &54.3\% \\
  DeepSeek-Math-7b (0-shot) & Unknown &PPO (QVM) \cite{ouyang2022training}          & Q* (QVM) \cite{lightman2023let} &\textbf{55.4}\% \\
 \midrule
   \end{tabular}}
   \label{tab_math}
\end{table*}

\textbf{MBPP.} As for the comparison on MBPP dataset, we also choose the most powerful open-source LLM in the aspect of code generation, specifically CodeQwen1.5-7b-Chat, as our base model for evaluating the effectiveness of Q*.
Following a similar procedure of math reasoning, we train a QVM for Q-value estimation and manually 
construct the utility function as described in the previous part of implementation details (tokenize the code generated so far with function \verb|tokenize.generate_tokens| and give a penalty of -0.5 if \verb|TokenError| is raised). 
From the results shown in Table \ref{tab_mbpp}, we can find that Q* can still outperform Best-of-$N$ method in the aspect of code generation, and help CodeQwen1.5-7b-Chat to achieve 77.0\% accuracy on MBPP dataset, which is also a promising performance in the leaderboard of MPBB \footnote{\url{https://paperswithcode.com/sota/code-generation-on-mbpp}}.

\begin{table*}
\caption{Comparison of Q* and other baselines on MBPP dataset.}
\renewcommand{\arraystretch}{1.25}
\centering
\scalebox{0.75}{
   \begin{tabular}{
   p{4.5cm}<{\centering}| p{3cm}<{\centering} | p{3cm}<{\centering} | p{3.5cm}<{\centering} | p{2.5cm}<{\centering}} 
   \toprule 
    \multicolumn{1}{c|}{Base Model} &\multicolumn{1}{c|}{SFT} &\multicolumn{1}{c|}{Alignment} &\multicolumn{1}{c|}{Verification} &\multicolumn{1}{c}{Accuracy}\\
    \midrule
    GPT-3.5 Turbo (self-debug) \cite{chen2023teaching} & Unknown & PPO (RM) \cite{ouyang2022training} & - & 72.8\% \\
    GPT-4 (self-debug) \cite{chen2023teaching}  & Unknown & PPO (RM) \cite{ouyang2022training} & - & \textbf{80.2}\% \\
   \midrule
    CodeQwen1.5-7b-Chat (0-shot) & Unknown &PPO (QVM) \cite{ouyang2022training}          & -&74.6\% \\
   CodeQwen1.5-7b-Chat (0-shot) & Unknown &PPO (QVM) \cite{ouyang2022training}          & Best-of-$N$ (QVM) \cite{lightman2023let} &75.0\% \\
  CodeQwen1.5-7b-Chat (0-shot) & Unknown &PPO (QVM) \cite{ouyang2022training}          & Q* (PRM+QVM) \cite{lightman2023let} &\textbf{77.0}\% \\
 \midrule
   \end{tabular}}
   \label{tab_mbpp}
\end{table*}

\section{Conclusion}
Solving challenging multi-step reasoning problems requires LLMs to perform in-depth deliberation beyond auto-regressive token generation. In this paper, we present Q*, a general, versatile and agile deliberation framework for LLMs. Unlike existing deliberation methods which need extensive expertise to design a utility function for each specific task, Q* relies on ground-truth solely to train value model and can be easily applied to various reasoning tasks without modification. Moreover, by leveraging plug-and-play Q-value models as the heuristic function, Q* can effectively guide LLMs to solve various tasks without fine-tuning LLMs beforehand, which avoids potential performance degeneration on other tasks. Finally, our Q* is agile because we consider only a single step each time rather than complete rollouts (\emph{e.g.}, simulation in MCTS). Extensive empirical evaluations on math reasoning and code generation tasks confirm the superiority of our method.
\bibliography{neurips_2024}
\bibliographystyle{unsrtnat}

\end{document}